\documentclass[runningheads]{llncs}
\usepackage[T1]{fontenc}
 
\usepackage{accvabbrv}
\usepackage{graphicx}
\usepackage{booktabs}
\usepackage{multirow}
\usepackage[utf8]{inputenc}
\DeclareUnicodeCharacter{2212}{-}
\usepackage{csquotes}
\usepackage{subcaption}
\usepackage{amsmath}
\usepackage{tabularx}
\usepackage{caption}
\usepackage{soul}
\usepackage{xspace}
\usepackage{xcolor}
\begin{document}

\title{LoLI-Street: Benchmarking Low-Light Image Enhancement and Beyond} 

\titlerunning{LoLI-Street: Benchmarking Low-Light Image Enhancement
and Beyond}

\author{Md Tanvir Islam\inst{1}\orcidID{0009-0007-9405-5684} \and
Inzamamul Alam \inst{1}\orcidID{0009-0004-6564-4867} \and 
Simon S. Woo\inst{1, *}\orcidID{0000-0002-8983-1542}\and
Saeed Anwar\inst{2}\orcidID{0000−0002−0692−8411}\and
IK Hyun Lee\inst{3}\orcidID{0000-0002-0605-7572}\and
Khan Muhammad\inst{3, *}\orcidID{0000-0002-5302-1150}}

\authorrunning{Md Tanvir Islam et al.}

\institute{Department of Software, Sungkyunkwan University, South Korea \and
The Australian National University, Canberra, Australia \and
Department of Mechatronics Engineering, Tech University of Korea, South Korea \and
Department of Human-AI Interaction, Sungkyunkwan University, South Korea \\
\email{*Corresponding authors:\{swoo, khanmuhammad\}@g.skku.edu}}

\maketitle
\begin{abstract}
Low-light image enhancement (LLIE) is essential for numerous computer vision tasks, including object detection, tracking, segmentation, and scene understanding. Despite substantial research on improving low-quality images captured in underexposed conditions, clear vision remains critical for autonomous vehicles, which often struggle with low-light scenarios, signifying the need for continuous research. However, paired datasets for LLIE are scarce, particularly for street scenes, limiting the development of robust LLIE methods. Despite using advanced transformers and/or diffusion-based models, current LLIE methods struggle in real-world low-light conditions and lack training on street-scene datasets, limiting their effectiveness for autonomous vehicles. To bridge these gaps, we introduce a new dataset \enquote{LoLI-Street} (Low-Light Images of Streets) with $33k$ paired low-light and well-exposed images from street scenes in developed cities, covering $19k$ object classes for object detection. LoLI-Street dataset also features 1,000 real low-light test images for testing LLIE models under real-life conditions. Furthermore, we propose a transformer and diffusion-based LLIE model named \enquote{TriFuse}. Leveraging the LoLI-Street dataset, we train and evaluate our TriFuse and SOTA models to benchmark on our dataset. Comparing various models, our dataset's generalization feasibility is evident in testing across different mainstream datasets by significantly enhancing images and object detection for practical applications in autonomous driving and surveillance systems.  Complete code and dataset is available on \textcolor{blue}{\textbf{https://github.com/tanvirnwu/TriFuse}}.

  \keywords{Low-light image enhancement \and LoLI-Street dataset\and \textcolor{black}{Conditional noise diffusion} \and Diffusion denoising \and Transformers}
\end{abstract}

\begin{figure}[t]
  \centering
  \begin{subfigure}{0.6\linewidth}

    \includegraphics[width=\textwidth]{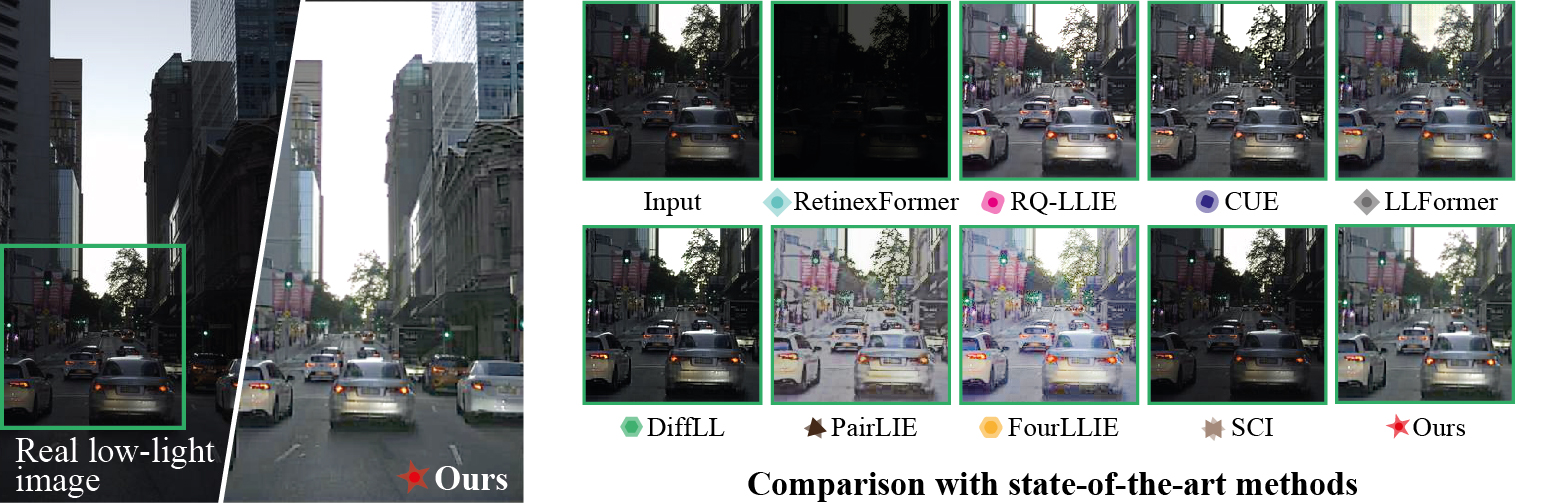}

    \caption{}
    \label{fig:fig1-a}
  \end{subfigure}
  \hfill
  \begin{subfigure}{0.35\linewidth}
 
    \includegraphics[width=\textwidth]{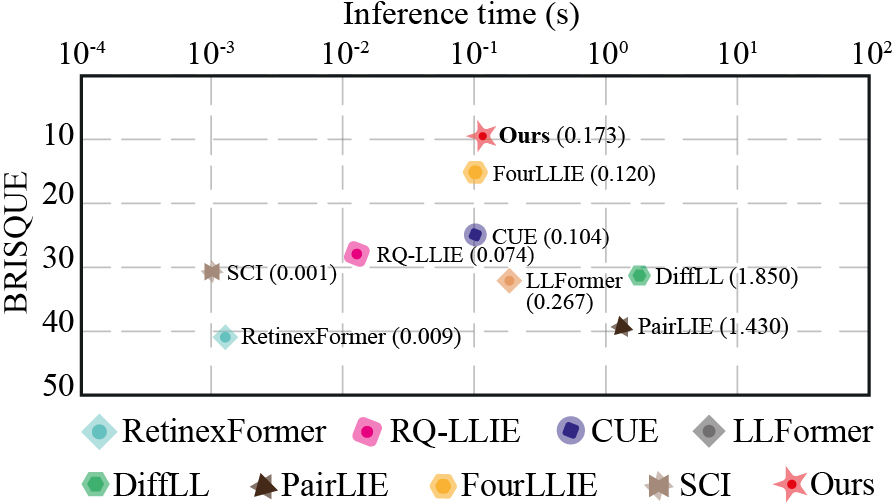}
  
    \caption{}
    \label{fig:fig1-b}
  \end{subfigure}

  \caption{Comparison between our TriFuse and SOTA models using a sample real low-light test image. (a) Qualitative comparison: Visually, PairLIE and RQ-LLIE produce brighter outputs but lack realism. In contrast, TriFuse ensures high visual quality with realistic enhancements. (b) Quantitative comparison based on the no-reference metric BRISQUE ($\downarrow$) and inference time ($\downarrow$).}
  \label{fig:fig1}

\end{figure}

\section{Introduction}
\label{sec:intro}

Low-light environments can pose significant challenges for various computer vision tasks in our daily lives. For most computer vision tasks, models are typically trained on datasets collected during the day with sufficient lighting, making them less effective in dark or low-light environments. This limitation poses a significant challenge as the underlying datasets do not account for the variations and complexities in real-world low-light conditions. Thus, as daylight fades into night, the reduced visibility can hinder the ability to perform even the most basic tasks for computer vision systems. This is a matter of convenience, safety, and efficiency. To address these practical challenges, advancements in computer vision technology are crucial. Such systems can significantly assist in low-light conditions, enhance the vision capabilities of autonomous vehicles~\cite{mandal2022real,li2021deep}, and improve safety and security measures~\cite{li2021deep}. The importance of computer vision in mitigating the effects of low-light conditions highlights its potential impact on a wide range of applications~\cite{szeliski2022computer}. For instance, recent advancements in image processing and machine learning have led to sophisticated algorithms that enhance image clarity~\cite{li2021low}, detecting~\cite{mittal2020deep} and recognizing~\cite{wang2022review} objects in near-darkness.

Additionally, with the rise of deep learning~\cite{panetta2022deep,xia2021deep,moran2020deeplpf}, transformers~\cite{wang2023ultra,lv2023unsupervised,cai2023retinexformer,wang2023ultra}, and diffusion methods~\cite{yin2023cle,jiang2023low,yi2023diff,wang2023exposurediffusion}, its feature representation capabilities led to the rapid adoption of LLIE. Moreover, researchers are exploring the latest transformer and diffusion-based methods for LLIE by utilizing synthetic datasets and reporting significant improvements in LLIE. However, the models struggle in practical applications, a huge gap that leaves the scope to develop robust methods for effective LLIE in real-world scenarios. Thus, the full potential of these methods for LLIE has not yet been fully explored and requires further research. Furthermore, these learning-based methods heavily rely on high-quality labeled data for training to perform accurately in real-world scenarios.

In literature, different datasets are available with different scene types of images under various low-light conditions~\cite{Chen2018Retinex,loh2019getting,chen2018learning,chen2024waterpairs,li2019underwater,duarte2016dataset}. Despite having several LLIE datasets, there is still a lack of datasets, especially for urban street image scene types, which can be used to train the LLIE models for autonomous vehicles to use navigation and surveillance cameras in urban street scenarios where accurate object detection, recognition, and navigation are crucial for safety.

To address these gaps in this paper, we contribute as follows:

\begin{itemize}
\item We introduce a unique and challenging dataset named LoLI-Street consisting of 30,000 train, 3,000 validation, and 1,000 real low-light test (RLLT) images featuring street scene types, which are rare among the existing datasets and feature three intense levels (high, moderate, and light) of low-light effect. 
\item We propose ``TriFuse'' that reduces the number of sampling steps in the diffusion process by using the transformer as an accurate noise predictor.
\item Benchmarking our proposed TriFuse method against SOTA LLIE models on LoLI-Street real low-light testset and mainstream datasets, we found it excels in LLIE and object detection, as shown in Fig. \ref{fig:fig1} and detailed in Section \ref{sec:comparison}.
\end{itemize}

\section{Related Works}

\subsection{LLIE Datasets} 
The ExDARK dataset~\cite{loh2019getting} includes 7,363 annotated images across 12 classes, crucial for low-light object detection. The LLVIP dataset~\cite{jia2021llvip} provides 15,488 pairs of visible and infrared images, essential for image fusion and pedestrian detection. The MIT-Adobe FiveK dataset~\cite{fivek} offers 5,000 indoor and outdoor images for various enhancement tasks. The SICE~\cite{cai2018learning} dataset synthesizes 589 images across varied illumination conditions, while the SID~\cite{chen2018learning} dataset pairs 5,094 short-exposure images with long-exposure references. Additionally, the LIME~\cite{guo2016lime} dataset features 10,000 images for LLIE in low-light conditions, and the DPED~\cite{ignatov2017dslr} dataset enhances mobile photo quality. The LOLv1~\cite{wei2018deep} and LOLv2~\cite{yang2020fidelity} datasets contain paired high and low-light images, and the LSRW dataset~\cite{hai2023r2rnet} includes paired low-light images. These datasets have indoor and outdoor scenes, as presented in Table~\ref{tab:main-a}. To the best of our knowledge, there is no dataset that presents the street scene types, unlike our proposed LoLI-Street dataset, which is crucial for autonomous vehicles under real-world low-light street scenarios. Moreover, our LoLI-Street provides a test set of 1000 street scene-type images under real-life low-light conditions for testing the LLIE methods.

\subsection{LLIE Methods}

\textbf{Transformer-based LLIE.}
Initially proposed for natural language processing~\cite{vaswani2017attention}, transformers have recently shown remarkable performance in computer vision tasks, such as image classification~\cite{ali2021xcit,arnab2021vivit,dosovitskiy2020image}, semantic segmentation~\cite{cao2022swin,wu2021visual,zheng2021rethinking}, and object detection~\cite{carion2020end,dai2021dynamic}. They have also proven effective in low-level vision tasks like image restoration~\cite{cai2023retinexformer,wang2023ultra} and image synthesis~\cite{hudson2021generative,jiang2021transgan,zhang2022styleswin}. Recent studies highlight transformers' effectiveness in LLIE by utilizing illumination-guided multi-head self-attention mechanisms to improve interactions between regions of different exposure levels~\cite{cai2023retinexformer}.

\subsubsection{Diffusion-based LLIE.}
Diffusion-based models have shown significant potential in LLIE by leveraging their generative capabilities to handle various degradations, including noise~\cite{zeng2024unmixing}, low contrast~\cite{lv20232}, color correction~\cite{cha2024descanning}, and medical-image denoising~\cite{gungor2023adaptive,ozbey2023unsupervised}. Recent advancements include Diff-Retinex~\cite{yi2023diff}, which combines Retinex with generative diffusion networks for enhanced detail and noise reduction. Integrating generative networks with physical models has led to effective restoration of scene structures~\cite{huang2023bootstrap}. Other recent research has introduced a conditional diffusion model incorporating multi-scale patch-based training~\cite{nguyen2024diffusion} and a wavelet-based conditional diffusion model~\cite{jiang2023low}, improving visual quality. Also, CLE Diffusion~\cite{yin2023cle} offers controllable light enhancement using classifier-free guidance, while ExposureDiffusion~\cite{wang2023exposurediffusion} combines a diffusion model with a physics-based exposure model for enhanced performance and reduced inference time. And, LDM integrates a denoising diffusion probabilistic model with a light enhancement network, achieving state-of-the-art performance for LLIE~\cite{lv20232}.

Despite recent advancements, existing LLIE methods struggle with real-world low-light images, especially on our challenging LoLI-Street dataset, making them unsuitable for autonomous vehicles and surveillance cameras. Our proposed model, TriFuse, leverages this street-scene dataset to achieve significant improvements over SOTA models testing on the RLLT.

\begin{figure}[!b]
  \centering
  \begin{subfigure}{0.6\linewidth}
    \includegraphics[width=\textwidth]{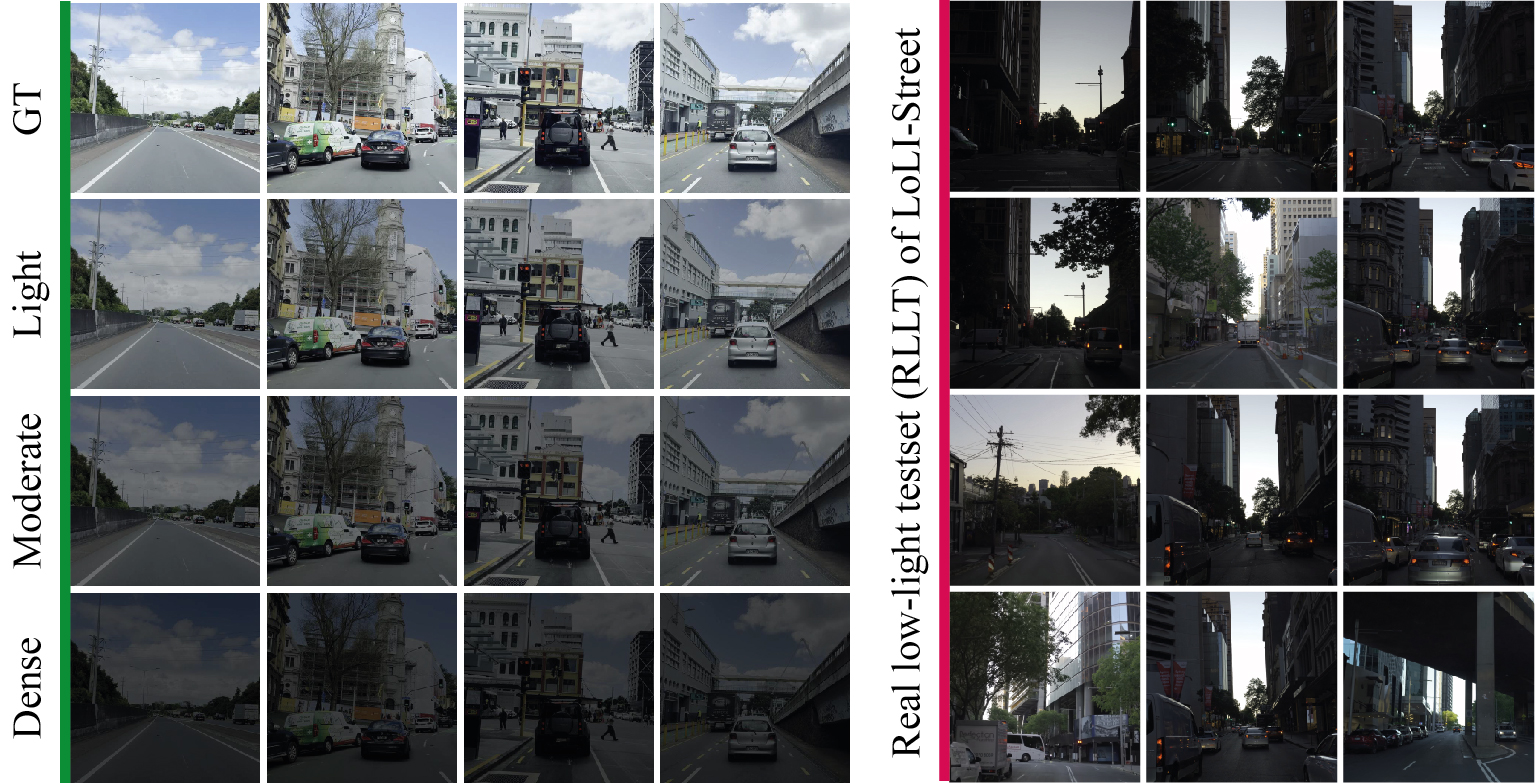}
    \caption{}
    \label{fig:dataA}
  \end{subfigure}
  \hfill
  \begin{subfigure}{0.35\linewidth}
    \includegraphics[width=\textwidth]{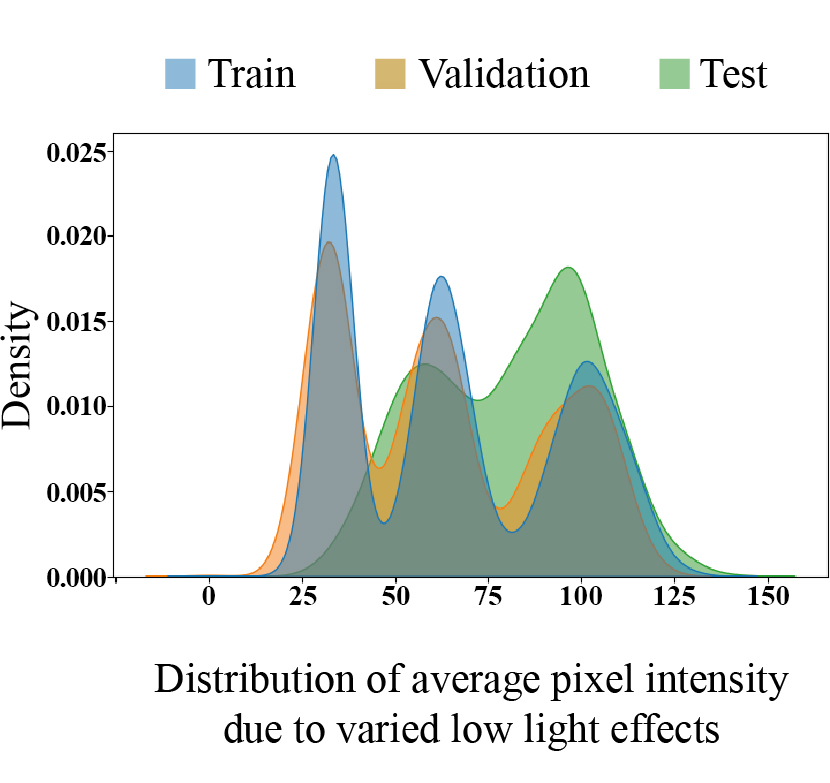}
    \caption{}
    \label{fig:dataB}
  \end{subfigure}
  \caption{(a) Sample images of LoLI-Street. \textbf{Green:} train and validation sets, \textbf{Red:} real low-light test set. (b) Distribution of the low-light images of our LoLI-Street.}
  \label{fig:short}
\end{figure}

\section{Methodology}

\subsection{Our Dataset: LoLI-Street}
We introduce the benchmark dataset ``Low-light Images of Streets (LoLI-Street)'', containing three subsets: train, validation, and test. The train and validation sets consist of $30k$ and $3k$ paired low and high-light images, and
the real low-light testset (RLLT) contains $1k$ images under real-world low-light conditions, totaling $33k$ images. We collected high-resolution videos (4K/8K at 60fps) from various cities under low-light conditions, extracting and manually reviewing frames to create the Real Low-light Testset (RLLT) of our LoLI-Street dataset, ensuring high quality and excluding any with motion blur. As shown in Table~\ref{tab:main-b}, LoLI-Street encompasses three levels of low-light intensity, resulting in different quantitative metrics. Sample images are presented in Fig.~\ref{fig:dataA}, and Fig.~\ref{fig:dataB} shows the average pixel distribution across subsets. Inspired by \cite{rshaze,haze1k}, we used Photoshop v25.0 to generate the synthetic images of our dataset and examined the distribution of the images. As evident from Fig.~\ref{fig:dataB}, the distribution of our dataset varied across the subsets, which is crucial for generalizing LLIE models.

\begin{table}[!t]
\caption{Quantatitive comparison of mainstream datasets and our LoLI-Street.}
\label{tab:main}
\centering \tiny
\hspace*{-1.5cm}
\begin{tabular}{cc}
    \begin{subtable}[t]{0.32\textwidth}
        \caption{Datasets comparison.}
        \label{tab:main-a}  
        \centering 
        \begin{tabular}{llc}
        \toprule
        \textbf{Dataset} & \multirow{2}{*}{\textbf{Venue}} & \textbf{\# of} \\ 
        \textbf{Name} &  & \textbf{Images}\\
        \midrule
        LOLv1~\cite{wei2018deep} & BMVC'18 & 1,500 \\
        SICE~\cite{Cai2018deep} & TIP'18   & 5,389 \\ 
        ExDARK~\cite{loh2019getting} & CVIU'19 & 7,363 \\
        LOLv2~\cite{lolv2} & CVPR'20 & 3,576 \\
        LLVIP~\cite{jia2021llvip} & ICCV'21 & 15,488 \\
        LSRW~\cite{hai2023r2rnet} & JVCIR'23 & 11,300 \\
        \midrule  
        \textbf{LoLI-Street} & \textbf{ACCV'24} & \textbf{34,000} \\
        \textbf{(Ours)} &  &  \\

      \bottomrule
  \end{tabular}
    \end{subtable}
    &
    \begin{subtable}[t]{0.60\textwidth}
        \caption{Quantitative analysis of our LoLI-Street dataset.}

        \label{tab:main-b}  
        \centering \tiny
        \begin{tabular}{@{}lcccccccccc@{}}
    \toprule
    \multirow{2}{*}{\textbf{Metrics}} && \multicolumn{3}{c}{\textbf{Train}} && \multicolumn{3}{c}{\textbf{Validation}} && \multirow{2}{*}{\textbf{Test}} \\
    \cline{3-5} \cline{7-9}
    && \textbf{Light} & \textbf{Moderate} & \textbf{Dense} && \textbf{Light} & \textbf{Moderate} & \textbf{Dense} &&\\
    \midrule
     PSNR↑ && 28.35 & 27.88 & 27.89 && 28.44 & 27.91 & 27.87 && - \\
    SSIM↑ && 0.8564 & 0.6045 & 0.3528 && 0.8767 & 0.6196 & 0.3398 && - \\
    MS-SSIM↑ && 0.9531 & 0.7943 & 0.5854 && 0.9422 & 0.7818 & 0.5621 && - \\
    LPIPS↓ && 0.0410 & 0.1547 & 0.2988 && 0.04199 & 0.1563 & 0.2490 && - \\
     MSE↓ && 106.14 & 95.71 & 105.99 && 106.79 & 94.77 & 108.02 && - \\
     MAE↓ && 204.51 & 166.09 & 137.21 && 206.09 & 169.68 & 142.15 && - \\
    \midrule
    
    BRISQUE↓ && 21.99 & 24.82 & 33.46 && 15.80 & 18.00 & 26.34 && 30.99 \\
    NIQE↓ && 11.045 & 12.119 & 13.352 && 10.49 & 10.49 & 10.49 && 12.334 \\
  \bottomrule
  \end{tabular}
    \end{subtable}
\end{tabular}
\end{table}

\subsection{Our Proposed Method}
Our proposed TriFuse integrates a custom vision transformer, wavelet-based conditional diffusion denoising, and an edge-sharpening module detailed as follows:

\subsubsection{Discrete Wavelet Transformation (DWT).}
\begin{figure}[!t]
  \centering
  \includegraphics[width=\textwidth]{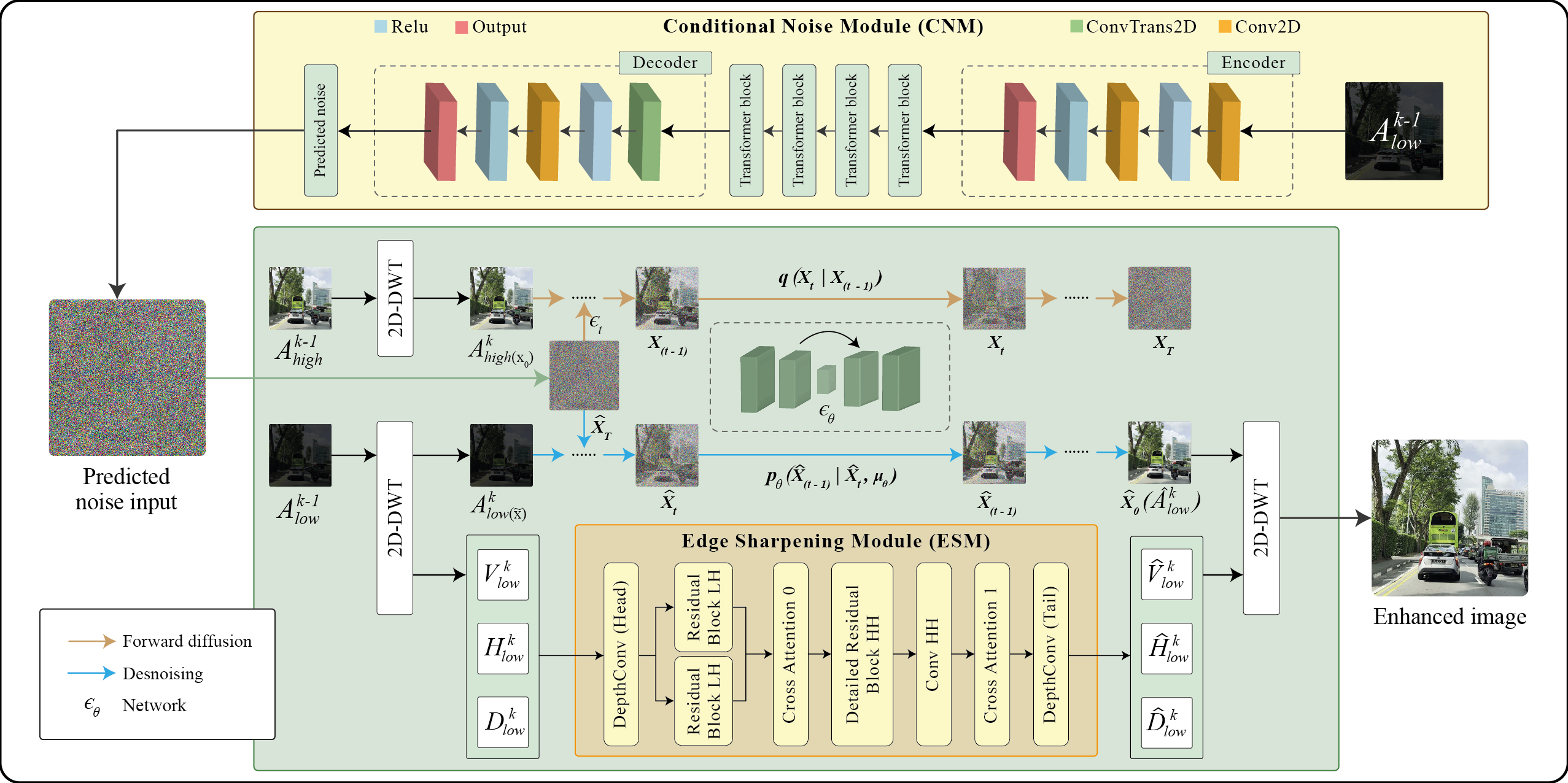}
  \caption{Overview of our TriFuse model, featuring the Conditional Noise Module (CNM) and Edge Sharpening Module (ESM) for effective LLIE. The CNM generates noise, refined through forward and backward passes by the diffusion process. The ESM sharpens the output image's edges. The process starts with predicted noise, undergoes wavelet transformation and denoising within TriFuse, and results in a visually enhanced image.}
  \label{fig:TriFuse-arch} 
\end{figure}

We use DWT to decompose a given low-light image $ I_{\text{low}} \in {R}^{H \times W \times C} $ in various low and high-frequency components. The 2D-DWT with Haar wavelets~\cite{haar1911theorie} decomposes the image into four sub-bands: $ A_1^{\text{low}} $, $ V_1^{\text{low}} $, $ H_1^{\text{low}} $, and $ D_1^{\text{low}} $, as illustrated in Fig.~\ref{fig:TriFuse-arch}. 
The mathematical formulation for the 2D-DWT is provided in Eq. (\ref{eq:eq1}):

\begin{equation}
\small
\label{eq:eq1}
\{A_1^{\text{low}}, V_1^{\text{low}}, H_1^{\text{low}}, D_1^{\text{low}}\} = \text{2D-DWT}(I_{\text{low}}),
\end{equation}

\noindent
where $ A_1^{\text{low}} $ is the approximation coefficient presenting the low-frequency information, and $ V_1^{\text{low}} $, $ H_1^{\text{low}} $, and $ D_1^{\text{low}} $ are the coefficients presenting the vertical, horizontal, and diagonal high-frequency information, respectively. Focusing the diffusion process on these components, especially the average coefficients, TriFuse enhances the model's ability to handle global image structures effectively.

\subsubsection{TriFuse.}
TriFuse integrates a transformer, CNN, Encoder, and Decoder block, involving the diffusion process for predicting noise at each timestamp, forming the cornerstone of our conditional noise generation for diffusion denoising. This approach leverages the power of transformers to accurately predict and adjust noise at each diffusion timestep of denoising diffusion probabilistic models (DDPM)~\cite{ho2020denoising}, enhancing the denoising process, which eventually improves LLIE.

In the forward diffusion process in Eq. (\ref{eq:eq2}), the input image $ x_0 $ is progressively corrupted into a noisy version $ x_T $ over $ T $ steps, governed by a variance schedule $\{\beta_1, \beta_2, \ldots, \beta_T\}$ as follows:

\begin{equation}
\small
\label{eq:eq2}
q(x_{1:T} | x_0) = \prod_{t=1}^{T} q(x_t | x_{t-1}), q(x_t | x_{t-1}) = \mathcal{N}(x_t; \sqrt{1 - \beta_t} x_{t-1}, \beta_t I),
\end{equation}

where $ X_t $ is noisy data at timestep $ t $, and $\beta_t$ is the variance schedule.

The reverse diffusion in Eq. (\ref{eq:eq3}) involves learning to denoise the noisy image $ x_T $ back to a clean image $ x_0 $ through a series of Gaussian denoising transitions:

\begin{equation}
\small
\label{eq:eq3}
p_{\theta}(\hat{x}_{0:T}) = p(\hat{x}_{T}) \prod_{t=1}^{T} p_{\theta}(\hat{x}_{t-1} \mid \hat{x}_{t}), \quad p_{\theta}(\hat{x}_{t-1} \mid \hat{x}_{t}) = \mathcal{N}(\hat{x}_{t-1}; \mu_{\theta}(\hat{x}_{t}, t), \sigma_t^2 \mathbf{I}).
\end{equation}

Here, $ \mu_{\theta} $ is the predicted mean, and $ \sigma_t $ is the variance, are learned parameters.

\subsubsection{Conditional Noise Module (CNM) for Diffusion Denoising.}
The CNM is designed to predict the noise $\epsilon_t$ at each timestep $t$, utilizing a transformer-based architecture to grasp the intricate patterns in noise and image details. Our model utilizes self-attention mechanisms to capture long-range dependencies and contextual information, unlike traditional diffusion models that rely on random Gaussian noise at each timestep. By conditioning the noise on the input image and the timestep, our CNM significantly enhances the denoising process.

The CNM architecture begins by encoding the input image into a higher-dimensional space using convolutional layers, which extract features. These encoded features are flattened and processed through a series of transformer blocks where the self-attention mechanism enables the model to assess the importance of different image parts, effectively predicting the noise to be added or removed. After transforming the features through self-attention and feed-forward layers, the output is reshaped to the original feature map dimensions and passed through a decoder, which reconstructs the predicted noise map, guides the diffusion.

The CNM's ability to model complex dependencies and incorporate contextual information results in superior image restoration, particularly in challenging low-light conditions. By accurately predicting and controlling the noise at each diffusion step, the CNM ensures an effective and precise denoising process, preserving fine details and maintaining contextual awareness.

\textcolor{black}{This integration enhances image quality by preserving fine details, contextual awareness, and providing adaptive denoising. Mathematically, the noise prediction is expressed as $\epsilon_\theta(\hat{x}_t, t)$ = $\text{CNM}(\hat{X}_T)$. After integrating our custom CNM with the process shown in Eq. (\ref{eq:eq3}), it can be expressed as Eq. (\ref{eq:eq4}) as follows:}

\begin{equation}
\small
\label{eq:eq4}
\hat{x}_{t-1} = \frac{1}{\sqrt{\alpha_t}} \left( x_t - \frac{1-\alpha_t}{\sqrt{1-\bar{\alpha}_t}} \text{CNM}(\hat{X}_T) \right) + \sigma_t \eta
\end{equation}
where $\alpha_t$ and $\bar{\alpha}_t$ are predefined noise schedules, and $\eta$ represents Gaussian noise.

\noindent
Overall, this novel approach ensures that the denoising process is both effective and precise by accurately predicting and controlling the noise at each diffusion step. The integration of the CNM enhances image quality by ensuring that the noise prediction is conditional on both the image content and the timestep, leading to superior restoration of image details in low-light conditions.

\subsubsection{Edge Sharpening Module (ESM).}
ESM plays a critical role in enhancing the sharpness and clarity of edges in the restored images. It focuses on the high-frequency components obtained from the DWT, ensuring that fine details and textures are well preserved during the restoration process.

The ESM comprises several sophisticated components designed to handle high-frequency information efficiently. Depthwise convolutions capture channel-wise spatial information effectively, ensuring that the model can focus on intricate details without increasing computational complexity. Dilated Residual Blocks ($\phi$) preserve the input's spatial resolution while capturing multi-scale features as provided in Eq. (\ref{eq:eq5}). Using dilated convolutions allows the network to have a larger receptive field, which is essential for capturing contextual information at multiple scales without losing fine details.

\begin{equation}
\small
\label{eq:eq5}
\mathbf{Y} = \mathbf{X} + \text{Conv}(\text{ReLU}(\text{BN}(\text{Conv}(\text{ReLU}(\text{BN}(\mathbf{X}))))))),
\end{equation}
\noindent
where $\mathbf{X}$ denotes the input feature map that enters the Dilated Residual Block, and $\mathbf{Y}$ is the output feature map after processing through the block. $\text{Conv}$, $\text{ReLU}$, and $\text{BN}$ denote convolution, Rectified Linear Unit, and Batch Normalization, respectively. Cross-attention mechanisms are used to align and integrate contextual information across different directions (vertical, horizontal, and diagonal). The cross-attention mechanism is defined in Eq.~(\ref{eq:eq6}) as follows:

\begin{equation}
\small
\label{eq:eq6}
\mathbf{A}_{\text{attn}} = \text{Softmax}\left( \frac{\mathbf{Q} \mathbf{K}^\top}{\sqrt{d_k}} \right) \mathbf{V},
\end{equation}

\noindent
where $ \mathbf{Q} = \text{Conv}(\mathbf{X}) $, $ \mathbf{K} = \text{Conv}(\mathbf{X}) $, $ \mathbf{V} = \text{Conv}(\mathbf{X}) $ are the query, key, and value matrices, and $ d_k $ is the dimensionality of the key vectors. The ESM processes the high-frequency components as given in Eq.~(\ref{eq:eq7}):

\begin{equation}
\small
\label{eq:eq7}
\text{ESM}(x) = x + \text{Conv}(\text{Concat}(\text{$\phi$}_{\text{HL}}(x_{\text{HL}}), \text{$\phi$}_{\text{LH}}(x_{\text{LH}}), \text{$\phi$}_{\text{HH}}(x_{\text{HH}}))),
\end{equation}
\noindent
where $ x_{\text{HL}}, x_{\text{LH}}, x_{\text{HH}} $ are the high-frequency components and $\text{$\phi$}_{\text{HL}}$, $\text{$\phi$}_{\text{LH}}$, $\text{$\phi$}_{\text{HH}}$ are the corresponding dilated residual blocks, respectively. By integrating these components, the ESM enhances the sharpness of edges and preserves the fine details in the restored images, addressing one of the critical challenges in LLIE.

Overall, our proposed TriFuse model produces high-quality, sharp images by combining the ESM and CNM modules in the diffusion denoising process, making it an efficient solution for LLIE and suitable for various real-world applications.

\begin{table}[!b]
  \caption{Quantitative evaluation of the SOTA models on the validation set of LoLI-Street dataset using the pre-trained weights of each model.}
  \label{tab:pretrained}
  \centering \tiny
  \resizebox{\textwidth}{!}{
  \begin{tabular}{lccccccccc}
    \toprule
    Models
    & RetinexF. \cite{cai2023retinexformer} 
    & RQ-LLIE \cite{liu2023low}
    & CUE \cite{zheng2023empowering} 
    & LLFormer \cite{wang2023ultra}  
    & DiffLL \cite{jiang2023low}
    & PairLIE \cite{fu2023learning}
    & FourLLIE \cite{wang2023fourllie}
    & SCI \cite{ma2022toward} 
    \\ \midrule

    Venue
    & ICCV'23 
    &ICCV'23  
    &ICCV'23 
    &AAAI'23 
    &TOG'23
    & CVPR'23 
    & ACM MM'23 
    & CVPR'22\\
    \midrule 
    
    Metrics & \multicolumn{8}{c}{Light}\\ \midrule 
    PSNR$\uparrow$ & 27.89 & 28.11 & 27.66 & \textbf{28.42} & 28.16 & \underline{28.17} & 28.06 & 27.62  \\
    SSIM$\uparrow$ & 0.1900 & 0.7382 & \underline{0.9000} & \textbf{0.9274} & 0.8932 & 0.8374 & 0.8363 & 0.7877 \\
    MS-SSIM$\uparrow$ & 0.4200 & 0.8506 & \underline{0.9331} & \textbf{0.9497} & 0.9148 & 0.8562 & 0.8196 & 0.7890 \\
    LPIPS$\downarrow$ & 0.4300 & 0.1637 & \underline{0.0603} & \textbf{0.0486} & 0.1069 & 0.1574 & 0.1996 & 0.2415 \\
    MSE$\downarrow$ & 110.33 & 101.30 & 111.58 & \textbf{94.49} & 99.48 & \underline{99.36} & 101.99 & 112.51  \\
    MAE$\downarrow$ & 93.01 & 93.01 & \textbf{45.36} & 86.41 & 114.12 & 70.74 & 87.12 & \underline{69.35}  \\

    \midrule

    &\multicolumn{8}{c}{Moderate}\\ \midrule 
    PSNR$\uparrow$ & 27.73 & 27.88 & \underline{28.02} & 27.99 & \textbf{28.16} & 28.00 & 27.96 & 27.81  \\
    SSIM$\uparrow$ & 0.0900 & 0.5145 & 0.6389 & \textbf{0.9018} & 0.8709 & 0.8651 & 0.5397 & \underline{0.8737} \\
    MS-SSIM$\uparrow$ & 0.3000 & 0.7552 & 0.7964 & \textbf{0.9468} & 0.8914 & 0.8764 & 0.8335 & \underline{0.9083}  \\
    LPIPS$\downarrow$ & 0.6000 & 0.2321 & 0.1606 & \textbf{0.0530} & 0.1365 & 0.1294 & 0.1831 & \underline{0.1021}  \\
    MSE$\downarrow$ & 106.87 & 106.20 & 103.98 & 104.02 & \textbf{99.41} & \underline{103.09} & 104.14 & 107.61  \\
    MAE$\downarrow$ & 124.52 & 124.51 & 85.66 & \textbf{44.76} & 119.65 & 101.05 & 113.81 & \underline{48.02}  \\

\midrule

    &\multicolumn{8}{c}{Dense}\\ \midrule 
    PSNR$\uparrow$ & 27.75 & 27.87 & 27.94 & \textbf{28.67} & \underline{28.56} & 28.06 & 28.07 & 28.16  \\
    SSIM$\uparrow$ & 0.0300 & 0.3498 & 0.3651 & \textbf{0.9056} & 0.8744 & \underline{0.8915} & 0.8828 & 0.8758  \\
    MS-SSIM$\uparrow$ & 0.2100 & 0.5987 & 0.5916 & \underline{0.9488} & 0.9322 & 0.8963 & 0.8780 & \textbf{0.9633} \\
    LPIPS$\downarrow$ & 0.7900 & 0.2889 & 0.2971 & \underline{0.0441} & 0.0795 & 0.1016 & 0.1187 & \textbf{0.0274} \\
    MSE$\downarrow$ & 110.89 & 106.29 & 104.52 & \textbf{89.01} & \underline{91.05} & 101.65 & 101.52 & 99.36  \\
    MAE$\downarrow$ & 116.85 & 116.85 & \underline{111.97} & \textbf{80.34} & 160.96 & 144.18 & 186.56 & 212.06 \\

  \bottomrule
  \end{tabular}}
\end{table}

\begin{table}[!th]
  \caption{Performance comparison of mainstream SOTA models and our proposed TriFuse on the LoLI-Street validation set using LoLI-Street-trained weights.}
  \label{tab:LoLI-Street-trained}
  \centering 
  \resizebox{\textwidth}{!}{
  \begin{tabular}{lcccccccccc}
    \toprule
    Methods 
    & RetinexF. \cite{cai2023retinexformer} 
    & RQ-LLIE \cite{liu2023low}
    & CUE \cite{zheng2023empowering} 
    & LLFormer \cite{wang2023ultra}  
    & DiffLL \cite{jiang2023low}
    & PairLIE \cite{fu2023learning}
    & FourLLIE \cite{wang2023fourllie}
    & SCI \cite{ma2022toward} 
    & TriFuse\\ \midrule

    Venue
    & ICCV'23 
    &ICCV'23  
    &ICCV'23 
    &AAAI'23 
    &TOG'23
    & CVPR'23 
    & ACM MM'23 
    & CVPR'22
    &(Ours)\\
    \midrule
    Metrics & \multicolumn{8}{c}{Light}\\ \midrule

    PSNR$\uparrow$  & 27.92 & 27.66 & 28.77 & \textbf{33.40} & 32.59 & 28.78 & 28.06 & 27.84 & \underline{32.89}\\
    SSIM$\uparrow$  & 0.8767 & 0.8811 & 0.6493 & \textbf{0.9648} & 0.9560 & 0.9169 & 0.8363 & 0.8759 & \underline{0.9585} \\
    MS-SSIM$\uparrow$  & 0.9422 & 0.9035 & 0.3177 & \underline{0.9876} & 0.9889 & 0.9372 & 0.8196 & 0.9413  & \textbf{0.9899}\\
    LPIPS$\downarrow$ & 0.0419 & 0.1038 & 0.4194 & \textbf{0.0039} & 0.0139 & 0.0625 & 0.1996 & 0.0429  & \underline{0.0107}\\
    MSE$\downarrow$ & 106.79 & 112.01 & 89.34 & \textbf{30.29} & 38.55 & 86.62 & 101.99 & 106.87  & \underline{34.06}\\
    MAE$\downarrow$ & 206.09 & \textbf{55.35} & 125.41 & 93.59 & 115.68 & \underline{84.86} & 87.12 & 206.12 & 107.63\\ 

    \midrule 
    &\multicolumn{8}{c}{Moderate} \\ \midrule 

    PSNR$\uparrow$  & 28.44 & 27.59 & 30.58 & \textbf{32.15} & \underline{31.87} & 28.42 & 27.96 & 28.38 & \textbf{32.15}\\
    SSIM$\uparrow$  & 0.6197 & 0.8829 & 0.9100 & \underline{0.9386} & 0.9352 & 0.9242 & 0.8693 & 0.6192 & \textbf{0.9462}\\
    MS-SSIM$\uparrow$  & 0.7819 & 0.9372 & 0.9668 & \textbf{0.9837} & 0.9789 & 0.9374 & 0.8336 & 0.7809 & \underline{0.9819}\\
    LPIPS$\downarrow$ & 0.1563 & 0.0668 & 0.0201 & \textbf{0.0061} & \underline{0.0142} & 0.0447 & 0.1831 & 0.1572 & \underline{0.0142}\\
    MSE$\downarrow$ & 94.78 & 113.49 & 57.41 & \textbf{40.14} & 40.38 & 93.75 & 104.14 & 94.88 & \underline{43.16}\\
    MAE$\downarrow$ & 169.68 & \textbf{44.33} & 148.72 & \underline{80.25} & 138.32 & 162.81 & 113.81 & 169.69  & 107.29\\
\midrule

    &\multicolumn{8}{c}{Dense}\\ \midrule 
    PSNR$\uparrow$ & 27.87 & 29.03 & 30.58 & \underline{31.62} & 31.04 & 27.66 & 28.07 & 27.79 & \textbf{31.67} \\
    SSIM$\uparrow$ & 0.3398 & 0.9167 & 0.9100 & \textbf{0.9274} & 0.9165 & 0.8702 & 0.8828 & 0.3394 & \underline{0.9214} \\
    MS-SSIM$\uparrow$ & 0.5621 & 0.9616 & 0.9668 & \textbf{0.9738} & \underline{0.9734} & 0.9413 & 0.8780 & 0.5614 & \underline{0.9734} \\
    LPIPS$\downarrow$ & 0.3037 & 0.0326 & 0.0201 & \textbf{0.0131} & 0.0273 & 0.0357 & 0.1188 & 0.3048 & \underline{0.0201} \\
    MSE$\downarrow$ & 108.02 & 82.51 & 57.41 & \underline{45.39} & 51.66 & 111.57 & 101.52 & 108.11 & \textbf{45.01} \\
    MAE$\downarrow$ & 142.15 & \underline{107.57} & 148.72 & 108.04 & 123.79 & 214.17 & 186.56 & 142.17 & \textbf{78.88} \\
    
  \bottomrule
  \end{tabular}}
\end{table}

\begin{table}[t]
  \caption{Quantitative comparison of mainstream SOTA models and our TriFuse on the LoLI-Street real low-light testset using LoLI-Street-trained weights.}
  \label{tab:real-LoLI-Street-trained}
  \centering 
  \resizebox{\textwidth}{!}{
  \begin{tabular}{@{}lccccccccc@{}}
    \toprule
    Models 
    & RetinexF. \cite{cai2023retinexformer} 
    & RQ-LLIE \cite{liu2023low}
    & CUE \cite{zheng2023empowering} 
    & LLFormer \cite{wang2023ultra}  
    & DiffLL \cite{jiang2023low}
    & PairLIE \cite{fu2023learning}
    & FourLLIE \cite{wang2023fourllie}
    & SCI \cite{ma2022toward} 
    & TriFuse\\ \midrule

    Metrics & \multicolumn{8}{c}{Pre-trained weights}\\ \midrule

    BRISQUE$\downarrow$ & 54.12 & 29.76 & \underline{13.65}  & \textbf{12.69} & 18.54 & 39.65 & 15.05 &  38.05&\multirow{2}{*}{-} \\
    NIQE$\downarrow$ & \textbf{11.79}& 	12.57 &	14.36	& 16.40 &	\underline{12.16} &	12.28 & 12.57 & \underline{12.16} & \\
    \midrule
    &\multicolumn{8}{c}{Trained weights}\\ \midrule
    BRISQUE$\downarrow$ & 41.69 & 29.76 & 25.97 & 30.44 & 30.11 & 35.25 & \underline{14.50} & 30.96 & \textbf{10.32}\\
    NIQE$\downarrow$ & 11.83& 12.57& 12.44& 11.84&	12.30&	\underline{11.78} & 11.89 & 12.32 & \textbf{10.61} \\
  \bottomrule 
  \end{tabular}}
\end{table}

\begin{table}[!t]
  \caption{Performance comparison between SOTA LLIE models and our proposed TriFuse testing on the mainstream LLIE datasets.}
  \label{tab:main-comp}
  \centering 
  \resizebox{\textwidth}{!}{
  \begin{tabular}{llccccccccc}
    \toprule
    \multirow{2}{*}{Dataset} & Models 
    & RetinexF. \cite{cai2023retinexformer} 
    & RQ-LLIE \cite{liu2023low}
    & CUE \cite{zheng2023empowering} 
    & LLFormer \cite{wang2023ultra}  
    & DiffLL \cite{jiang2023low}
    & PairLIE \cite{fu2023learning}
    & FourLLIE \cite{wang2023fourllie}
    & SCI \cite{ma2022toward} 
    & TriFuse\\ \cline{2-11}

    \cline{2-11}
    &Metrics & \multicolumn{8}{c}{Full-reference metrics}\\
    \midrule
    \multirow{6}{*}{\rotatebox{90}{\textbf{LOLv1}}} & PSNR$\uparrow$ & 27.89 & \underline{28.00} & 27.97 & 27.77 & 27.88 & 27.82 & 27.93 & 27.95 & \textbf{28.01} \\
    & SSIM$\uparrow$ & 0.6299 & 0.8181 & \underline{0.8724} & 0.7778 & 0.8207 & 0.7111 & 0.7074 & 0.2333 & \textbf{0.8756} \\
    & MS-SSIM$\uparrow$ & 0.7542 & \textbf{0.8642} & 0.8562 & 0.8389 & 0.8555 & 0.8313 & 0.7932 & 0.4956 & \underline{0.8578} \\
    & LPIPS$\downarrow$ & 0.2072 & \textbf{0.1157} & 0.1418 & 0.1502 & 0.1473 & 0.1448 & 0.1595 & 0.4177 & \underline{0.1410} \\
    & MSE$\downarrow$ & 106.44 & 105.14 & \underline{104.59} & 109.08 & 103.02 & 108.18 & 105.25 & 104.42 & \textbf{102.83} \\
    & MAE$\downarrow$ & 174.08 & \underline{172.88} & 176.54 & 179.50 & 169.74 & 179.70 & 182.58 & 150.58 & \textbf{145.05} \\

\midrule
    \multirow{6}{*}{\rotatebox{90}{\textbf{LOLv2 (R)}}}& PSNR$\uparrow$ & 27.82 & 27.82 & \textbf{28.13} & 27.72 & \underline{27.88} & 27.77 & 27.62 & 27.73 & \underline{27.88} \\
    & SSIM$\uparrow$ & 0.5650 & 0.7799 & \underline{0.8437} & 0.7670 & 0.7875 & 0.7246 & 0.7473 & 0.2543 & \textbf{0.8966} \\
    & MS-SSIM$\uparrow$ & 0.3261 & 0.8753 & \textbf{0.8951} & 0.8651 & 0.8695 & 0.8612 & 0.8167 & 0.5542 & \underline{0.8823} \\
    & LPIPS$\downarrow$ & 0.5583 & \underline{0.0988} & 0.1057 & 0.1163 & 0.1219 & 0.1021 & 0.1580 & 0.3518 & \textbf{0.0888} \\
    & MSE$\downarrow$ & 107.39 & 109.26 & \textbf{102.06} & 110.77 & 108.32 & 109.78 & 112.71 & 109.94 & \underline{107.32} \\
    & MAE$\downarrow$ & \underline{170.48} & 180.82 & 174.28 & 185.34 & 175.12 & 199.54 & 205.84 & 176.08 & \textbf{170.28} \\
    
 \midrule

    \multirow{6}{*}{\rotatebox{90}{\textbf{LOLv2 (S)}}} & PSNR$\uparrow$ & 28.02 & 28.07 & 28.01 & 28.03 & 28.03 & 28.14 & \textbf{29.61} & 28.03 & \underline{28.70}\\
    & SSIM$\uparrow$ & 0.6511 & 0.6392 & 0.6516 & 0.6759 & 0.5967 & 0.8223 & \textbf{0.9623} & 0.4857 & \underline{0.8593}\\
    & MS-SSIM$\uparrow$ & 0.7756 & 0.7741 & 0.7905 & 0.7799 & 0.7644 & \underline{0.8623} & \textbf{0.9656} & 0.7059 & 0.8236\\
    & LPIPS$\downarrow$ & 0.2449 & 0.2470 & 0.2397 & 0.2374 & 0.2567 & 0.1697 & \textbf{0.0403} & 0.2969 & \underline{0.0727}\\
    & MSE$\downarrow$ & 103.06 & 102.03 & 103.26 & 103.01 & 102.85 & 101.06 & \textbf{76.78} & 103.17 & \underline{98.48}\\
    & MAE$\downarrow$ & 158.50 & 164.45 & 159.11 & \underline{156.94} & 159.62 & 167.85 & \textbf{121.24} & 176.12 & 175.73\\

\midrule
    \multirow{6}{*}{\rotatebox{90}{\textbf{LSRW}}} & PSNR$\uparrow$ & \underline{27.96} & 28.00 & 27.93 & 27.95 & 27.98 & 27.94 & 27.98 & 27.81 & \textbf{28.03}\\
    & SSIM$\uparrow$ & 0.6160 & 0.6590 & 0.6751 & 0.6543 & 0.6099 & 0.6684 & \underline{0.7298} & 0.2938 & \textbf{0.7459}\\
    & MS-SSIM$\uparrow$ & 0.6832 & 0.6900 & 0.6959 & 0.6958 & 0.6846 & \textbf{0.7045} & 0.6970 & 0.5399 & \underline{0.6988}\\
    & LPIPS$\downarrow$ & 0.1873 & 0.1648 & 0.1592 & 0.1685 & 0.1805 & \underline{0.1358} & 0.1748 & 0.3526 & \textbf{0.1350}\\
    & MSE$\downarrow$ & 104.56 & 103.52 & \underline{102.69} & 104.77 & 104.10 & 105.12 & 103.77 & 107.88 & \textbf{100.14}\\
    & MAE$\downarrow$ & 182.63 & 176.41 & 177.46 & 182.88 & 172.70 & 185.33 & 174.83 & \underline{172.28} & \textbf{152.92}\\

\midrule

    \multirow{6}{*}{\rotatebox{90}{\textbf{SICE}}} & PSNR$\uparrow$ & 28.02 & 27.96 & 27.97 & 28.02 & 27.97 & \textbf{28.34} & \underline{28.14} & 27.81 & 27.92 \\
    & SSIM$\uparrow$ & 0.8078 & 0.7681 & 0.7605 & \underline{0.8092} & 0.7530 & \textbf{0.8843} & 0.8796 & 0.6582 & 0.8001\\
    & MS-SSIM$\uparrow$ & 0.8396 & 0.8275 & 0.8279 & \underline{0.8397} & 0.8333 & 0.8462 & \textbf{0.8739} & 0.8112 & 0.8240\\
    & LPIPS$\downarrow$ & 0.1537 & 0.1654 & 0.1679 & \underline{0.1519} & 0.1606 & 0.1252 & \textbf{0.1082} & 0.1774 & 0.1587\\
    & MSE$\downarrow$ & 102.89 & 104.24 & 104.15 & 102.85 & 104.02 & \textbf{96.10} & \underline{101.22} & 107.85 & 102.18 \\
    & MAE$\downarrow$ & 157.14 & 163.18 & 155.44 & 156.13 & 161.02 & \textbf{116.67} & 122.22 & 183.48 & \underline{118.72}\\

    \midrule
    &&\multicolumn{9}{c}{No-reference metrics}\\
    \midrule
    \multirow{2}{*}{\textbf{ExDark}} & BRISQUE$\downarrow$ & 22.90 & 33.56 & 17.81 & \textbf{16.92} & 22.19 & 32.29 & 18.08 & 38.95 & \underline{17.29}\\
    & NIQE$\downarrow$ & 13.93 & 15.83 & \textbf{10.40} & 16.250 & 13.56 & 14.93 & 13.844 & 14.392 & \underline{13.47}\\ \midrule

    \multirow{2}{*}{\textbf{LLVIP}} 
    & BRISQUE$\downarrow$ & 25.09 & 26.23 & \underline{17.81} & 19.86 & 18.32& 34.99 & 21.55 & 23.65 & \textbf{10.32}\\
    & NIQE$\downarrow$ & \underline{10.66} & 10.79 &  17.22 & 10.62 &  11.37 & 10.96 & \textbf{10.64} &  11.78 & 11.07\\

    \bottomrule 
  \end{tabular}}

\end{table}

\section{Experimental Setup}

\subsubsection{Datasets.}
We use the train set of $30k$ paired images from our LoLI-Street dataset to train the models and validate the models' performance on the synthetic validation set of $3k$ paired images. Furthermore, we evaluate the models on the real test set of the LoLI-Street dataset, including $1k$ unpaired images. We used the well-known LOLv1 and LOLv2 datasets to evaluate the pre-trained and trained weights of each model and compare the existing models' performance with our TriFuse model. From LOLv1, we use the validation set, which features 15 paired real low-light images to evaluate the models. Similarly, from LOLv2, we take the synthetic and real validation subsets of 100 paired images from each to assess the models. The LSRW dataset features two paired subsets of Huawei and Nikon camera-captured images, which we combine to get 50 paired images to evaluate the models' performance. ExDark and LLVIP unpaired datasets check the models' effectiveness under highly dark conditions. 

\subsubsection{Implementation.} We used PyTorch on a server with four NVIDIA RTX 2080 GPUs (24GB each). SOTA models were trained with default settings for fair comparison. Our TriFuse was trained with a batch size of 12 and a patch size of 256$\times$256. The initial learning rate of $1 \times 10^{-4}$ decayed by 0.8 every $5 \times 10^3$ iterations. For efficient restoration, the time step $T$ was set to 200, and the implicit sampling step $S$ was set to 5 for both the training and inference phases.

\subsubsection{Evaluation Strategy.}
We calculated full-reference metrics (PSNR, SSIM, MS-SSIM, MSE, and MAE) and no-reference metrics (BRISQUE, and NIQE) to evaluate existing models and our TriFuse model. We assessed SOTA LLIE models with pre-trained weights on our LoLI-Street dataset to evaluate their quality. Additionally, we trained these models on our dataset and tested them to determine its suitability for generalization. Finally, we quantitatively and qualitatively compared our proposed model with recent SOTA models.

\begin{table}[!t]
  \caption{Computational complexity comparison of proposed TriFuse and the SOTA LLIE models. \textbf{AIT:} Average inference time ($\downarrow$).}
  \label{tab:complexity}
  \centering \scriptsize
  \begin{tabular}{lccccccccc}

    \toprule
    Models 
    & RetinexF. 
    & RQ-LLIE 
    & CUE 
    & LLF.
    & DiffLL 
    & PairLIE 
    & FourLLIE
    & SCI 
    & TriFuse (Ours)\\ \midrule

    Params  & 1.61M & 11.38M & 0.25M & 24.55M & 20.09M &0.34M & \underline{0.12M} & \textbf{384} & 26.48M\\
    FLOPS & \underline{15.57G} & 162.07G & 157.32G &39.61G & 89.60G & 89.38G & 1.95G & \textbf{0.14G} & 147.03G\\
    AIT & \underline{0.009s} & 0.074s  & 0.104s & 0.267s & 1.850s & 1.4304s & 0.120s& \textbf{0.001s} & 0.173s\\
    
    \bottomrule 
  \end{tabular}
\end{table}

\section{Comparative Analysis}
\label{sec:comparison}
We compare our TriFuse with multiple SOTA LLIE methods, including RetinexFormer~\cite{cai2023retinexformer}, RQ-LLIE~\cite{liu2023low}, CUE~\cite{zheng2023empowering}, LLFormer~\cite{wang2023ultra}, DiffLL~\cite{jiang2023low}, PairLIE~\cite{fu2023learning}, FourLLIE~\cite{wang2023fourllie}, and SCI~\cite{ma2022toward}, covering transformer and diffusion-based models.

\begin{table}[!b]
  \caption{Object detection performance using YOLOV10 on LoLI-Street after LLIE using different models. \textbf{MC:} Motorcycle, \textbf{TL:} Traffic light, \textbf{SS:} Stop sign.}
  \label{tab:object-detection}
  \centering 
  \resizebox{\textwidth}{!}{
  \begin{tabular}{@{}lccccccccc@{}}
    \toprule
    Models
    & RetinexF. \cite{cai2023retinexformer} 
    & RQ-LLIE \cite{liu2023low}
    & CUE \cite{zheng2023empowering} 
    & LLF. \cite{wang2023ultra}  
    & DiffLL \cite{jiang2023low}
    & PairLIE \cite{fu2023learning}
    & FourLLIE \cite{wang2023fourllie}
    & SCI \cite{ma2022toward} 
    & TriFuse\\ \midrule

    Metrics & \multicolumn{9}{c}{Average of mAP for various IoU thresholds for object detection after enhancing images using each model}\\ \midrule

    mAP(0.5) $\uparrow$ & 0.548 & 0.602 & 0.586 &  0.623 & \underline{0.650} & 0.548 & 0.483 & \underline{0.650} & \textbf{0.753} \\ 
    mAP(0.5-0.9) $\uparrow$ & 0.476 & 0.521 & 0.510 & 0.562 & 0.568 & 0.476 & 0.388 & \underline{0.653} & \textbf{0.692}\\ \midrule

    \multicolumn{10}{c}{mAP for varying IoU(0.5-0.9)  values detecting some important classes after enhancing images using each model}\\ \midrule
    
    Person  & 0.762 & 0.723 & 0.736 & \underline{0.779} & 0.749 & 0.737 & 0.586 & 0.760 & \textbf{0.791}\\
    Bicycle  & 0.600 &  0.508 & 0.584 & 0.601 & 0.570 & 0.541 & 0.405 & \textbf{0.660} & \underline{0.650} \\
    Car &  0.876 & 0.854 & 0.858 & \textbf{0.906} & 0.870 &  0.858 & 0.782 & 0.884 & \underline{0.891}\\
    MC & 0.641 &  0.618 & 0.646 & 0.648 &  0.649 & 0.642 & 0.540 & \underline{0.743} & \textbf{0.750}\\ 
    Bus  &  0.793 & 0.722 & 0.716 &  0.797 & \underline{0.825} & 0.714 & 0.616 & 0.820 & \textbf{0.851}\\
    TL & 0.666 & 0.580 & 0.642 & \underline{0.695} & 0.621 & 0.596 & 0.356 & 0.652 & \textbf{0.852}\\
    SS  & 0.447 &  0.343 & 0.458 & 0.518 & 0.635 & 0.496 & 0.206 & \textbf{0.821} & \underline{0.785}\\

    \bottomrule 
  \end{tabular}}
\end{table}

\subsubsection{Quantitative Analysis.}
We present a quantitative analysis of SOTA models on the LoLI-Street and existing datasets. Table~\ref{tab:pretrained} shows the performance of these models against the validation set using pre-trained weights with full-reference metrics under various lighting conditions. LLFormer performs robustly across all subsets, achieving the highest PSNR of 28.67 for the dense variety of our validation set. Table~\ref{tab:LoLI-Street-trained} evaluates SOTA models on the LoLI-Street validation set using LoLI-Street-trained weights, showing significant performance improvements and model generalization. Our proposed TriFuse achieves the highest scores in various metrics, demonstrating its robustness and effectiveness in LLIE tasks.

The performance of the SOTA models is presented in Table~\ref{tab:real-LoLI-Street-trained} on the real low-light test set of LoLI-Street, using both pre-trained and trained weights for each model. The evaluation metrics include BRISQUE and NIQE. Our proposed model, TriFuse, stands out with the lowest BRISQUE and NIQE scores, indicating superior visual quality and naturalness of the enhanced images compared to the existing models. Table~\ref{tab:main-comp} provides a performance comparison of the SOTA models and our proposed TriFuse on existing datasets (LOLv1, LOLv2 (real), LOLv2 (synthetic), LSRW, SICE, ExDark, and LLVIP). As shown, we observed that our model consistently achieved either the best or second-best performance across multiple datasets, as indicated by both full-reference and no-reference metrics. This further validates the effectiveness of our model and emphasizes its ability to generalize well from the training dataset. Table~\ref{tab:complexity} summarizes the computational complexity, demonstrating our model's balance between efficiency and performance with competitive FLOPS and inference time metrics. Overall, the quantitative analysis establishes that our proposed TriFuse model consistently outperforms existing SOTA models across various metrics and datasets, proving its effectiveness and robustness for LLIE tasks. 

Moreover, Table~\ref{tab:object-detection} presents object detection results on the validation set where TriFuse achieves the highest mAP(0.5) and mAP(0.5-0.9) values, where 0.5 indicates the Intersection over Union (IoU) threshold and 0.5-0.9 represents the average mAP over multiple IoU thresholds.

\begin{figure}[!ht]
  \centering
  \includegraphics[width=\textwidth]{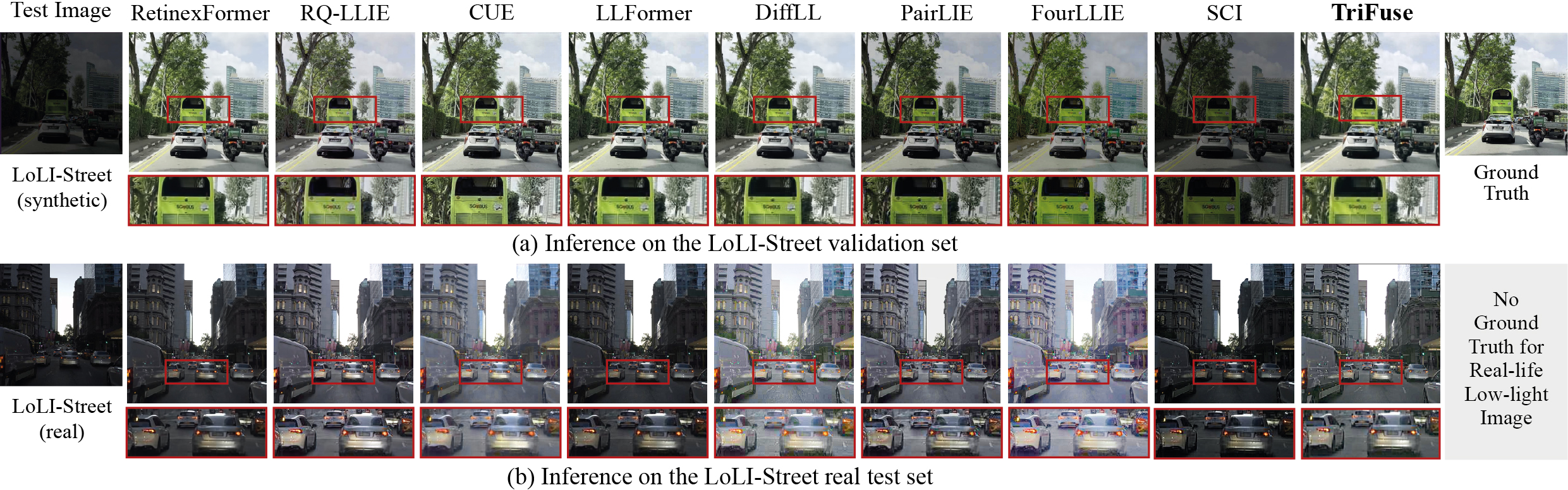}
  \caption{Enhanced images by models picking a random image from the (a) synthetic validation set and (b) real low-light test set of our LoLI-Street dataset.}  
  \label{fig:fig4}
\end{figure}

\begin{figure}[!th]
  \centering
  \includegraphics[width=\textwidth]{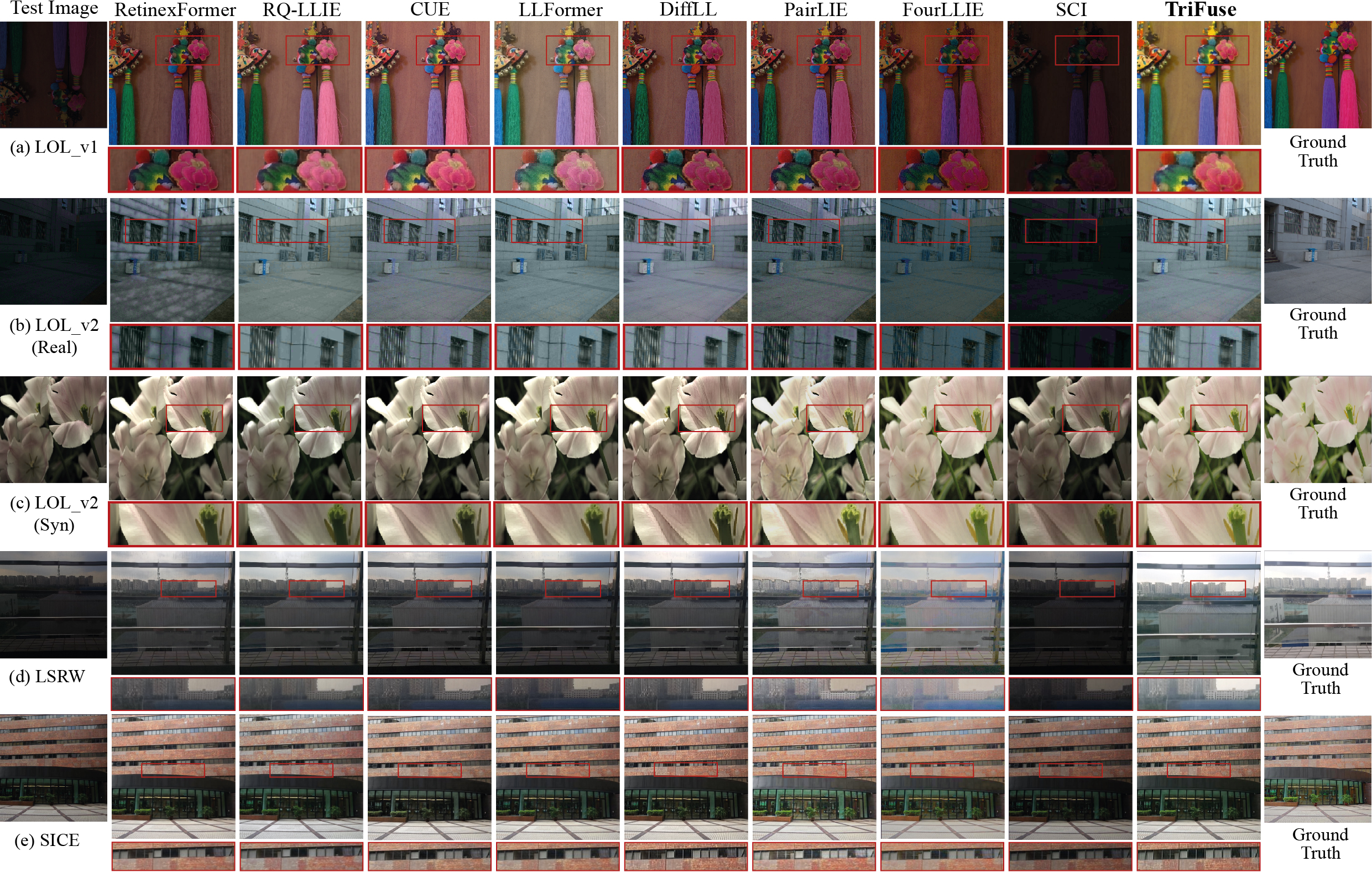}
  \caption{Enhanced images by SOTA models and our proposed TriFuse picking a random image from the validation sets of mainstream LLIE datasets.}  
  \label{fig:existingdata}

\end{figure}

\subsubsection{Qualitative Analysis.}
\begin{figure}[!th]
  \centering
  \includegraphics[width=\textwidth]{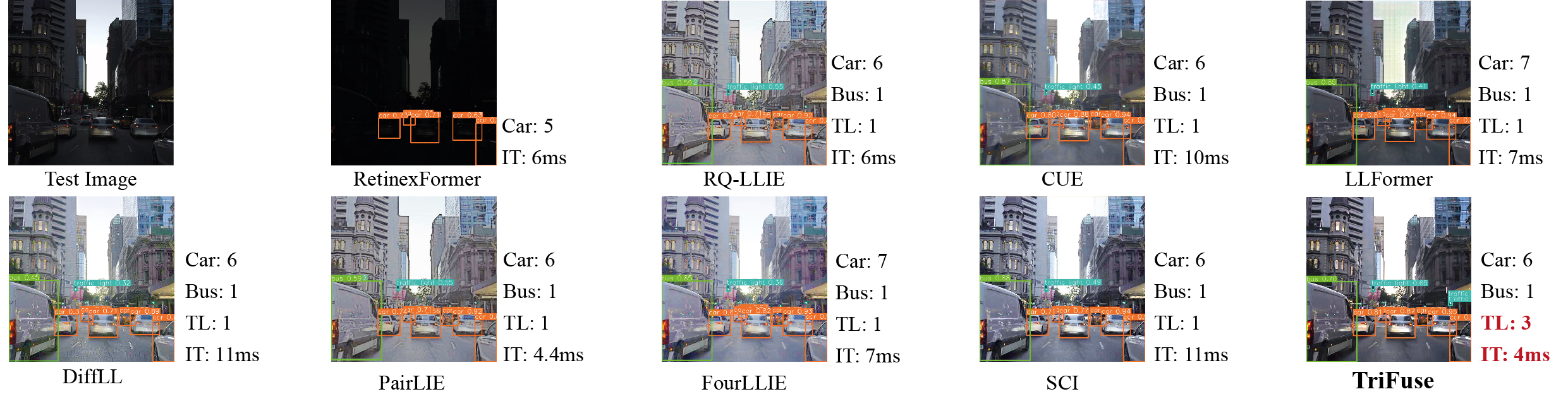}
  \caption{The outcome of YOLOV10 inference on a sample RLLT image of LoLI-Street dataset after enhancement. \textbf{TL:} Traffic light, \textbf{IT:} Inference time ($\downarrow$).}  
  \label{fig:fig7}
\end{figure}

In addition to the quantitative analysis, we conducted a qualitative evaluation of the enhanced images produced by different models on various datasets. Figure~\ref{fig:fig4} showcases enhanced images from the LoLI-Street dataset's synthetic validation set and real low-light test set, demonstrating that our model consistently provides clearer and more detailed visual enhancements, especially in shadowed and low-light areas. Figure~\ref{fig:existingdata} presents enhanced images from the LOLv1 and LOLv2 (both real and synthetic), LSRW, and SICE validation sets, where our model excels in color fidelity and enhancing image details, as evident in the close-up views, revealing well-maintained texture details and reduced artifacts. Overall, the comparison highlights TriFuse's robustness and superior performance in enhancing low-light images across multiple datasets. Also, Fig.~\ref{fig:fig7} illustrates the results of YOLOv10 inference on a randomly selected image from the LoLI-Street test set after enhancement by different models. Our model improves visual quality and enhances object detection accuracy, detecting additional objects such as traffic lights and cars with faster inference times compared to other approaches. This qualitative analysis demonstrates our model's effectiveness in enhancing low-light images, significantly improving visual quality and object detection performance in real-world conditions.

\subsubsection{Ablation Study.}
We perform a set of experiments as an ablation study with various combinations of components as presented in Table \ref{tab:ablation}. For the wavelet transformation scale, we compared the default setting $k(1)$ with $k(2)$ and $k(3)$. The results demonstrate that the ESM+ CNM+ k(1)+ S(5) configuration achieves superior BRISQUE and NIQE scores of 10.32 and 10.61, respectively, on the RLLT dataset, indicating enhanced visual quality compared to other settings. Evaluating the importance of ESM and CNM, comparisons with configurations excluding these components (w/o-ESM and w/o-CNM) highlight the superior performance of the default TriFuse setup. Analysis of different sampling steps (S(5), S(10), S(15)) reveals that increasing to S(15) enhances performance, achieving the highest PSNR of 33.37 and SSIM of 0.9470 on the validation set. Nevertheless, S(5) maintains competitive performance with superior computational efficiency and clarity in real low-light conditions, achieving the lowest BRISQUE (10.32) and NIQE (10.61) scores on the RLLT among all.

\begin{table}[!th]
  \caption{Ablation studies performed on TriFuse model with varying wavelet parameter, component, and diffusion sampling step. The underlined TriFuse represents the default setting (ESM+ CNM+ $k(1)$+ $S(5)$) of our model.}
  \label{tab:ablation}
  \centering \scriptsize
  \begin{tabularx}{\textwidth}{@{}l|l|c|c|X|X|X|X|X@{}}
    \toprule
    \multirow{2}{*}{Dataset} & \multirow{2}{*}{Metrics} & \multicolumn{2}{c|}{Wavlet Scale} & \multicolumn{3}{c|}{Components} & \multicolumn{2}{c}{Sampling Steps}\\ \cline{3-9}
     && $k(2)$ & $k(3)$ & w/o-ESM & w/o-CNM & \underline{TriFuse} & $S(10)$ & $S(15)$ \\ 
    \midrule
    
     \multirow{2}{*}{RLLT} & BRISQUE$\downarrow$  & 10.87 & 11.69 & 11.03 & 11.75 & \textbf{10.32} & 10.65  & 11.14 \\
     &NIQE$\downarrow$  & 11.25 & 11.85 &  11.58 & 12.01 & \textbf{10.61} & 10.95 & 11.52 \\
    \midrule

     Synthetic & PSNR$\uparrow$ & 31.02 & 31.63 & 31.74 & 30.87 & \textbf{32.24} & 32.88  & 33.37 \\
     Validation& SSIM$\uparrow$ & 0.9369 & 0.9312 &  0.9364 & 0.9227 & \textbf{0.9420} & 0.9411 & 0.9470\\
     
    \bottomrule 
  \end{tabularx}
\end{table}

\section{Conclusion}
Identifying the growing need for LLIE solutions, we introduced LoLI-Street, a novel benchmark dataset featuring street scenes under diverse lighting conditions designed to enhance images and improve object detection in low-light environments, which is crucial for autonomous systems. Our proposed LLIE model, TriFuse, incorporates a unique wavelet-based CNM approach to generate accurate input noise in the diffusion denoising process. This results in effective denoising for real-world LLIE in lower diffusion sampling steps. Comprehensive evaluations demonstrate TriFuse's superiority over existing SOTA models across multiple benchmarks, achieving top performance in visual quality and object detection under low-light conditions based on various metrics. Future directions include optimizing TriFuse for real-time applications and adapting it to diverse adverse scenarios~\cite{jin2023enhancing,hazespace2m} employing unsupervised techniques~\cite{jin2022unsupervised}.

\section{Acknowledgments}
This work was partly supported by Institute for Information \& communication Technology Planning \& evaluation (IITP) grants funded by the Korean government MSIT: (RS\ -2022-II221199, RS-2024-00337703, RS-2022-II220688, RS\ -2019-II190421, RS-2023-00230337, RS-2024-00356293, RS-2022-II221045, RS-2021-II212068, No.RS-2023-00231200, RS-2024-00437849). Lastly, this work was supported by the Korea Internet \& Security Agency (KISA) grant funded by the Korean government (PIPC) (No. RS-2023-00231200, Development of personal video information privacy protection technology capable of AI learning in an autonomous driving environment).

\bibliographystyle{splncs04}
\bibliography{0429}
\end{document}